\documentclass{article}
\usepackage{diagbox}
     \PassOptionsToPackage{numbers, compress}{natbib}
     \usepackage[final]{neurips_2019}

\usepackage[utf8]{inputenc} % allow utf-8 input
\usepackage[T1]{fontenc}    % use 8-bit T1 fonts
\usepackage{hyperref}       % hyperlinks
\usepackage{url}            % simple URL typesetting
\usepackage{booktabs}       % professional-quality tables
\usepackage{amsfonts}       % blackboard math symbols
\usepackage{nicefrac}       % compact symbols for 1/2, etc.
\usepackage{microtype}      % microtypography

\title{Towards Robust and Stable Deep Learning Algorithms for Forward Backward Stochastic Differential Equations}

\author{%
  Batuhan G\"{u}ler, Alexis Laignelet, Panos Parpas  \\
  Department of Computing\\
  Imperial College London\\
  London SW7 2AZ, UK \\
  \texttt{\{bbg2218, acl18, panos.parpas\}@imperial.ac.uk} \\
}

\usepackage{graphicx}
\usepackage{pgfplots}
\usepackage{amsmath}
\pgfplotsset{compat=1.14}
\DeclareMathOperator{\Tr}{Tr}
\usetikzlibrary{arrows}
\usepackage{verbatim}
\usepackage{mwe}
\usepackage{subfig}
\usepackage{makecell}

\begin{document}

\maketitle

\begin{abstract}
Applications in quantitative finance such as optimal trade execution, risk management of options, and optimal asset allocation involve the solution of high dimensional and nonlinear Partial Differential Equations (PDEs). The connection between PDEs and systems of Forward-Backward Stochastic Differential Equations (FBSDEs) enables the use of advanced simulation techniques to be applied even in the high dimensional setting. Unfortunately, when the underlying application contains nonlinear terms, then classical methods both for simulation and numerical methods for PDEs suffer from the curse of dimensionality. 
Inspired by the success of deep learning, several researchers have recently proposed to address the solution of FBSDEs using deep learning.  We discuss the dynamical systems point of view of deep learning and compare several architectures in terms of stability, generalization, and robustness. In order to speed up the computations, we propose to use a multilevel discretization technique. Our preliminary results suggest that the multilevel discretization method improves solutions times by an order of magnitude compared to existing methods without sacrificing stability or robustness. 
\end{abstract}

\section{Introduction}
The solution of nonlinear, high dimensional Partial Differential Equations (PDEs) is a long-standing problem throughout science and engineering. PDEs are also ubiquitous in quantitative finance where they appear in applications such as derivative pricing\cite{shreve2004stochastic}, optimal trade execution\cite{cartea2015algorithmic}, and optimal asset allocation\cite{bjork2009arbitrage}. Classical techniques such as finite difference and finite element methods suffer from the curse of dimensionality, i.e., the computational resources required to solve the problem increase exponentially with the number of dimensions. Due to the curse of dimensionality, it is only possible to solve PDEs in low dimensions (typically less than five). However, the PDEs that appear in many applications, and in finance in particular, are in much higher dimensional spaces. For example, pricing and hedging contracts for basket options, optimal trade execution, and asset allocation can have hundreds of dimensions. 

Pricing, hedging, and risk management in high dimensions is still an open problem in quantitative finance.  Recently researchers have proposed to use deep learning methods to solve high-dimensional PDEs (see \cite{han2018solving,henry2017deep,raissi2018forward,weinan2017deep}).
The proposed methods take advantage of the links between certain classes of PDEs and systems of Forward Backward Stochastic Differential Equations (see e.g. \cite{gobet2016monte}).
These proposals are motivated by the success of deep learning in areas of computer science such as computer vision and natural language processing. However, both theoretical and empirical results have shown that neural networks compute classifiers that are unstable
\cite{Moosavi_Dezfooli_2017, szegedy2013intriguing}. 
An unstable classifier is vulnerable to adversarial attacks and illegal exploitation\cite{szegedy2013intriguing}. The perturbations needed to fool ML classifiers are small, indistinguishable from noise, and therefore they are difficult to detect\cite{fawzi2018adversarial,szegedy2013intriguing,serban2018adversarial}. The lack of stability is not just a property of classifiers trained with neural networks. It also affects other algorithms such as kernel methods and support vector machines \cite{DBLP:journals/corr/PapernotMG16}. A necessary condition for successful ML systems in real-world applications, especially ones in the financial sector, is that the underlying system is stable. Without resolving this challenging problem, it is not possible to make meaningful progress in critical application areas such as the explainability and interpretability of machine learning algorithms, or efficient and robust training methods for reinforcement learning.

In this paper, we investigate the stability and robustness of deep learning for solving high dimensional PDEs. We argue that it is essential to develop both stable architectures as well as stable optimization algorithms. To address the stability of the network architecture, we adopt the recently proposed dynamical systems view of neural networks. The stability of the optimization algorithm to hyperparameters, such as the learning rate, can be addressed using proximal techniques that are reminiscent of implicit methods for solving differential equations. Due to space limitations we only focus on stable architectures in this paper. We will report the results on stable optimization algorithms in a separate paper.

The rest of the paper is structured as follows: In Section 2, we review the application of deep learning to high dimensional PDE and their relation to systems of Forward-Backward Stochastic Differential Equations (FBSDEs). In Section 3, we describe stable architectures for deep learning. Our stability considerations are based on classical discretization techniques as well as projection methods. We also discuss how to take advantage of multiple levels of discretization of the system of FBSDEs to speed up the computations. In Section 4 we report preliminary numerical results that show that the proposed multilevel approach can significantly reduce computational times without losing stability.

\section{Background: Using Neural Networks to Solve BSDE}\label{sec:background}
%Connection/Problem Set-up PDE/SDE/Neural Nets
In this section, we briefly review the links between certain classes of PDE and FBSDE. These links are quite standard and well understood, see for example \cite{gobet2016monte} for the general case and \cite{guyon2013nonlinear} for applications of FBSDEs to finance. Due to space limitations we refer the interested reader to the recent references \cite{han2018solving,henry2017deep,raissi2018forward,weinan2017deep} for a literature review regarding the formulation of PDE problems as learning problems. Below we only review the basics mainly to establish notation, and we omit many technical details.

\subsection{Links between PDE and SDE}
For $t\in[0,T]$ consider the following system of stochastic differential equations,
\begin{equation}
\label{eq:coupled} 
\begin{aligned}
	& dX_t = \mu(t, X_t, Y_t, Z_t) dt + \sigma(t, X_t, Y_t) dW_t, 
	 \ \ \ \ \ \ \  \ X_0 = \xi \\
	& dY_t = \varphi(t, X_t, Y_t, Z_t) dt + Z_t^T \sigma(t, X_t, Y_t)dW_t, 
	\ \ \ \ Y_T = g(X_T)
\end{aligned}
\end{equation}
where $X_0 = \xi\in\mathrm{R}^d$ is the initial condition for the forward SDE, and $Y_T = g(X_T)$ is the terminal condition of the backward SDE. $W_t$ is a vector valued Brownian motion. We assume that $g:\mathrm{R}^d\rightarrow \mathrm{R}$ is known. Regarding the data of the problem, i.e. $\mu, \sigma$ and $\varphi$ we make the same assumptions as Chapter 7 in \cite{gobet2016monte}. The solution of the system in \eqref{eq:coupled} is the triplet $\{X_t,Y_t,Z_t\}$. From a computational point of view the forward equation is easy to solve (using for example a simple Euler discretization). However the backward equation requires different solution techniques in order to ensure the solution is adapted to the filtration generated by the stochastic process in \eqref{eq:coupled}.

Consider the following non-linear PDE,
\begin{equation}
    u_t = f(t, x, u, Du, D^2u) \label{eq:general pde}\\
\end{equation}
where $Du, D^2u$ represent the gradient and Hessian of $u$ respectively. When the function $f$ is given by,
\begin{equation}
    f(t, x, y, z, \gamma) = \varphi (t, x, y, z) - \mu (t, x, y, z)^T z - \frac{1}{2} \Tr[\sigma(t, x, y) \sigma(t, x, y)^T \gamma],
\end{equation}
and the terminal condition of the PDE is given by $u(T, x) = g(x)$ then it is known that (e.g. Theorem 7.1 in \cite{gobet2016monte}), 
\begin{equation}
\begin{aligned}
	Y_t &= u(t, X_t) \\
	Z_t &= \nabla u(t, X_t). 
\end{aligned}
\end{equation}
Therefore the solution to the PDE can be obtained by solving \eqref{eq:coupled}. 
Note that the setup described above is for the case when the PDE is linear in $D^2u$ (the PDE is called semi-linear in this case). Similar results can be obtained for the case where $f$ is nonlinear in $D^2u$. The benchmark problems we consider in Section \ref{sec:numerical results} are for the semi-linear case.

\subsection{Solving FBSDEs using Deep Learning}
In the section above we described how the partial differential equation in \eqref{eq:general pde} is related to the forward-backward stochastic differential equations in \eqref{eq:coupled}. In this section we discuss the numerical solution of \eqref{eq:coupled} using a neural network. The formulation we use is similar to the one proposed in \cite{raissi2018forward}. 

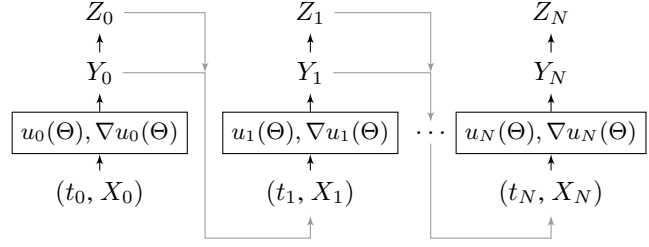
\begin{figure}[t]
\centering
\tikzstyle{line} = [draw, -latex', black]  
\tikzstyle{line_bis} = [draw, -latex', black!40] 
\tikzstyle{process} = [rectangle, fill=white, text centered, draw=black, font=\fontsize{9}{30}\selectfont]  
\tikzstyle{term} = [rectangle, fill=white, text centered]

\begin{tikzpicture}[node distance=0.8 cm, auto]  
    
    \node [term] (out0) {$Y_0$};  
    \node [term, above of=out0, node distance=0.8cm] (z0) {$Z_0$}; 
    \node [process, below of=out0] (ll0) {$u_0(\Theta),\nabla u_0(\Theta)$};  
    \node [term, below of=ll0] (in0) {$(t_0$, $X_0)$};
    
    \path [line] (out0) -- (z0); 
    \path [line] (ll0) -- (out0);  
    \path [line] (in0) -- (ll0);   

    \node [term, right of=out0, node distance=2.8cm] (out1) {$Y_1$};
    \node [term, above of=out1, node distance=0.8cm] (z1) {$Z_1$}; 
    \node [process, below of=out1] (ll1) {$u_1(\Theta),\nabla u_1(\Theta)$};  
    \node [term, below of=ll1] (in1) {$(t_1$, $X_1)$};    
    
    \path [line] (out1) -- (z1);     
    \path [line] (ll1) -- (out1);  
    \path [line] (in1) -- (ll1);
    
    \node [right of=ll1, node distance=1.6cm] (hl) {$\hdots$};

    \node [term, right of=out1, node distance=3.2cm] (outN) {$Y_N$};  
    \node [process, below of=outN] (llN) {$u_N(\Theta),\nabla u_N(\Theta)$}; 
    \node [term, above of=outN, node distance=0.8cm] (zN) {$Z_N$}; 
    \node [term, below of=llN] (inN) {$(t_N$, $X_N)$};  

    \path [line] (outN) -- (zN); 
    \path [line] (llN) -- (outN);  
    \path [line] (inN) -- (llN);  
    
    \draw [line_bis] (z0) -- ++(1.4,0) -- ++(0, -0.8);
    \draw [line_bis] (z1) -- ++(1.6,0) -- ++(0, -0.8);

    \draw [line_bis] (out0) -- ++(1.4,0) -- ++(0,-2.2) -- ++(1.4,0) -- (in1);
    \draw [line_bis] (out1) -- ++(1.6,0) -- (hl);
    \draw [line_bis] (hl) -- ++(0, -1.4) -- ++(1.6,0) -- (inN);

\end{tikzpicture}
  \caption{Approximation of a system of FBSDE using the same parametric approximation for every time step.}
  \label{fig:neural}
\end{figure}

If the solution of the PDE in \eqref{eq:general pde} was known, then we could apply a simple Euler-Maruyama scheme to approximate the solution to \eqref{eq:coupled}, i.e.
\begin{equation}\label{eq:disc equations}
\begin{aligned}
\Delta W_n  & \sim \mathcal{N}(0,\,\Delta t_n) \\
    X_{n+1} & \approx  X_n + \mu(t_n, X_n, Y_n, Z_n) \Delta t_n + \sigma(t_n, X_n, Y_n) \Delta W_n \\
    Y_{n+1} & \approx Y_n + \varphi(t_n, X_n, Y_n, Z_n) \Delta t_n + Z_n^T \sigma(t_n, X_n, Y_n) \Delta W_n \end{aligned}
\end{equation}
Where $Y_n=u(t_n,X_n), \ Z_n=\nabla u(t_n,X_n)$ and $\Delta t_n$ is a discretization parameter.
Since the solution of the PDE is not known then we parameterize the solution $u_t$ using a neural network with parameters $\Theta$ and compute $\nabla u$ using automatic differentiation, i.e. $u(t,x)\approx u(t,x;\Theta)$ and $\nabla u(t,x)\approx \nabla u(t,x;\Theta)$. The main idea is explained in Figure \ref{fig:neural}. Given an initial condition $X_0$ we use the parameterized solution to compute $(Y_t,Z_t)$ for $t\in[0,T]$. Note that we use the same parameters for all time points. Of course, we could use different parameters at every time point but from a computational point of view it is much better to use the same parameters. Our numerical experiments also suggest that it is sufficient to use a single set of parameters. Initially, the approximation obtained from the neural network is unlikely to satisfy the discretized equation in \eqref{eq:disc equations}. This observation leads to the following optimization problem,
\begin{equation}
\begin{split}
\label{eq:loss}
\min_{\Theta} \sum_{m=1}^{M} \sum_{n=0}^{N-1} &|Y_{n+1}^m(\Theta) - Y_n^m(\Theta) - \varphi(t_n, X_n^m, Y_n^m(\Theta), Z_n^m(\Theta)) \Delta t_n  \\
&- (Z_n^m(\Theta))^T \sigma(t_n, X_n^m, Y_n^m(\Theta)) \Delta W_n^m |^2 +
\sum_{m=1}^M \left|Y_N^m(\Theta) - g(X_N^m)\right| ^2
\end{split}
\end{equation}
where $M$ is the batch-size. In related works (\cite{han2018solving} and \cite{weinan2017deep}), the loss function only uses the terminal condition, and the architecture is different. In \cite{han2018solving} and \cite{weinan2017deep}, there are $N$ different neural networks. $Y_0$ and $Z_0$ are treated as parameters and all the $Y_n$ are computed using an Euler discretization. The loss function above simply compares the values obtain by the Euler scheme with the prediction obtained by the neural network. 
Assuming the function $g$ is differentiable, we could also add the following term to the loss function: $\sum_{m=1}^M \left|Z_N^m(\Theta) - g'(X_N^m)\right| ^2$.

%For each time step, the true value of $Y_{n+1}$ is the one given by the discretization. For the last time step $N$, the true value is also given by the terminal condition: $Y_T = g(X_T)$. It leads to this loss function, with $M$ the batch size:

%The benefit of making the comparison at each time step is heavily discussed in \cite{raissi2018forward}. Here, a unique neural network is used for every time steps, according to Figure \ref{fig:neural}. From the input $(t_n, X_n)$, the neural network produces $Y_n$. $Z_n$ is then computed using automatic differentiation from deep learning libraries.

\section{Stability and Generalisation}

\subsection{Neural networks as dynamical systems}
% Batu
Conventional deep neural networks struggle to backpropagate the gradient efficiently. Without heuristic methods and extensive hyper-parameter tuning, deep neural networks suffer from the exploding and vanishing gradient problems\cite{haber2017stable,li2017maximum}. As a result, a large number of heuristic techniques are needed to ensure gradient descent methods can train deep neural networks. Residual Neural Networks (Resnet) are a class of neural networks that attempt to address some of the shortcomings of conventional deep networks. Residual networks address the problem of having a large number of layers by performing identity mappings (shortcut connections). These identity mappings encourage the network to learn residual corrections as opposed to direct mappings. Resnet was proposed in 2015 and achieved state-of-the-art performance for deep neural networks in image recognition (\cite{he2016deep}).

Residual networks are defined as follows,
\begin{equation}
    x(k+1) = x(k) + f(x(k), \theta (k)),
\end{equation}
where $x(k)$ and $x(k+1)$ are the output of the $k^{th}$ layer and the $(k+1)^{th}$ layer respectively. The function $f$ denotes a non-linear transformation and $\theta (k)$ represents the parameters of the the $k^{th}$ layer. This relation can be seen as a forward Euler discretization of an ordinary differential equation with a step-size $h=1$ (\cite{haber2017stable}),
\begin{equation}\label{eq:ode}
    \dot{x} = f(x(t), \theta (t)). 
\end{equation}
This simple observation has allowed several authors to view the training of a neural network as an optimal control problem \cite{li2017maximum,haber2017stable,benning2019deep,liu2019deep,chen2018neural,haber2019imexnet}. An advantage of adopting the dynamical systems view of deep learning is that it allows the use of stable discretization techniques to the ODE in \eqref{eq:ode}. This idea has been explored by a number of authors (see \cite{benning2019deep} for a recent review). 

Residual networks address various issues, but they introduce the problem of forward stability (\cite{haber2019imexnet}). This refers to the amplification of an input perturbation during the forward pass. In this paper we use the non-autonomous network architecture first proposed in \cite{ciccone2018nais} called NAIS-Net. The reason we use the NAIS-Net model is that the resulting network is globally asymptotically stable for every initial condition. This is a crucial property for the PDE problem described in Section \ref{sec:background}.  NAIS-Net, a non-autonomous input-output stable neural network, tackles this issue by constraining the network as follows:
\begin{equation}
    x(k+1) = x(k) + h\sigma (Ax(k)+Bu+C)
\end{equation}
In this setting, A, B and C are trainable parameters. Parameter A refers to the traditional weight matrix and C refers to the bias. The extra term $Bu$ is made of a matrix $B$ and the input of the network $u$. Involving the input $u$ makes the system non-autonomous, and the output of the system input-dependent. Moreover, the weight matrix is constrained to be symmetric definite negative:
\begin{equation}
    A = -R^{T}R-\epsilon I
\end{equation}
where $\epsilon$ is a hyper-parameter that ensures the eigenvalues are strictly negative, $0<\epsilon$ ($\epsilon = 0.01$ in our case). 
An additional constraint is proposed on the Frobenius norm $\Vert R^{T}R \Vert_{F}$ with the algorithm. The projection used in this setting forces the weights of the neural network to stay within the set of feasible solutions and makes it more robust. We refer the interested reader to \cite{ciccone2018nais} for a detailed description of NAIS-Net. In our numerical experiments we found that NAIS-Net has the desired stability properties but is much slower than conventional Resnet architectures. Below we discuss an effective method to speed up the convergence of the algorithm that is inspired by Multilevel Monte Carlo Methods (MLMC).

\subsection{Convergence speed-up}\label{sec:mlmc}
The discretization parameter used to derive the approximation in \eqref{eq:disc equations} directly determines how many times the optimization algorithm needs to perform a forward solve to determine the loss function and backward propagation to determine the gradient of the loss (see Figure \ref{fig:neural}). Therefore the discretization parameter ($\Delta t$) has a dramatic effect on the computational cost of the method. We take advantage of the fact that we can vary the discretization parameter in order to speed-up the computations. 
The idea is inspired by Multilevel Monte Carlo Methods (ML-MC) that was proposed to reduce the variance of simulations with SDEs \cite{giles2008multilevel}. 
Instead of computing the discretization with a small step size we use a geometrically increasing step size. During the generation of our samples, we start with samples of small number of time steps (low accuracy) and then gradually increase the number of time steps. The main idea is that we take a small number of samples with a high level of accuracy and computational cost.

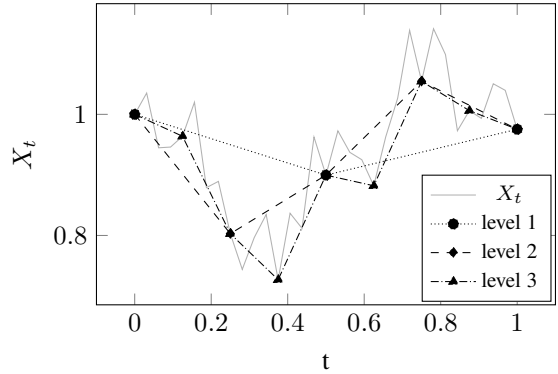
\begin{figure}[ht] 
\centering
\begin{tikzpicture}
\begin{axis}[xlabel={\makecell{t}},ylabel={$X_{t}$},width=0.55\textwidth, height= 0.4\textwidth, legend style={at={(0.85,0.01)},anchor=south},   legend style={font=\fontsize{8}{5}\selectfont}]

\addplot[color=black!30,dash pattern=]
table[x index=0,y index=1,col sep=comma] {data/levels.dat};

\addlegendentry{$X_t$}

\addplot[color=black,dash pattern=on \pgflinewidth off 1pt, mark=*]
table[x index=0,y index=1,col sep=comma] {data/level1.dat};

\addlegendentry{level 1}

\addplot[color=black,dash pattern=on 3pt off 3pt, mark=diamond*]
table[x index=0,y index=1,col sep=comma] {data/level2.dat};

\addlegendentry{level 2}

\addplot[color=black, dash pattern=on 3pt off 1pt on \the\pgflinewidth off 1pt, mark=triangle*]
table[x index=0,y index=1,col sep=comma] {data/level3.dat};

\addlegendentry{level 3}

\end{axis}
\end{tikzpicture}
  \caption{Simulation of a Geometric Brownian Motion for different levels of accuracy. The grey line represents the continuous-time stochastic process.}
  \label{fig:multilevel}  
\end{figure}

More formally, a SDE with general drift $a$ and volatility $b$, can be expressed as,
\begin{equation}
    dX_{t} = a(X_{t},t)dt+b(X_{t},t)dW_{t}.
\end{equation}
The discretization, with a time step $h$ gives:
\begin{equation}
    X_{n+1} = X_{n}+ a(X_{n},t_{n})h+b(X_{n},t_{n})\Delta W_{n}.
\end{equation}
Decreasing the time step $h$ will reduce the approximation error but the computational cost will increase. In the beginning, the error is large and the computational cost is small. When the error decreases, a finer grid is used with a higher computational cost. For instance, we can use:
$    h_{l} = h_{0}M^{-l}$
where $h_{0}$ is the initial step size, $M$ is a constant and $l$ is the index of the level. The number of time steps will increase exponentially over the levels. Figure \ref{fig:multilevel} illustrates the ML-MC scheme for an asset following a geometric Brownian motion. 

%%% Multilevel figure
%%% End multilevel figure

\section{Numerical results}\label{sec:numerical results}
In this section we discuss the numerical performance of the proposed method.
We performed numerical experiments on the following benchmark test problems.
\begin{itemize}
    \item Black-Scholes Equation:
    \[
    u_t=-\frac12 \text{Tr}[\sigma^2 \text{diag}(X^2)D^2u]+r(u-(Du)^\top x)
    \]
    The equation above is for the case when the underlying assets are uncorrelated and have the same variance $\sigma^2$. We used the boundary condition $g(x)=\|x\|^2$. In this case the Black-Scholes equation has the closed form solution,
    \[
    u(x,t)=e^{(r+\sigma^2)(T-t)}\|x\|^2.
    \]
    The experiments below are for the case when $d=100, \ r=0.05$ and $\sigma=0.4$.
    \item Hamilton-Jacobi-Bellman Equation:
    \[
    u_t=-\text{Tr}[D^2u]+\|D^2u\|^2, \ u(T,x)=g(x).
    \]
    The equation above has the closed form solution,
    \[
    u(t,x)=-\ln{\mathrm E[\exp(-g(x+\sqrt2 W_{t-t})}].
    \]
    In our experiments we used $g(x)=\ln(0.5(1+\|x\|^2), \ T=1, \ \sigma=\sqrt2$ and $d=100$.
    \item Allen-Cahn Equation,
    \[
    u_t=-0.5\text{Tr}[D^2u]-u+u^3, \ u(T,x)=g(x),
    \]
We set $g(x)=(2+0.4\|x\|)^{-1}$ and $d=20$.
\end{itemize}
We chose the test problems above because they have been used in several other recent works \cite{raissi2018forward,weinan2017deep,han2018solving}. The first two problems have a closed form solution and therefore it is easy to validate our implementation (see Figure \ref{fig:BS_results} for a comparison between the closed form solution and the one computed from the neural network). 
Due to space limitations we only report results from the first benchmark problem (Black-Scholes equation). The results for the other two test problems are similar and can be found here: \url{https://github.com/batuhanguler/Deep-BSDE-Solver}.

\subsection{Experimental setup}
The simulations are conducted with a batch size $M=100$, over $N=50$ time steps (except for the multi-level method described in Section \ref{sec:mlmc}). We use a neural network with 4 fully connected layers and $256$ neurons to approximate the solution of the PDE (see \url{https://github.com/batuhanguler/Deep-BSDE-Solver} for the exact details). The PyTorch implementation runs on a Tesla T4 GPU. We used the Adam optimizer and a learning rate of $10^{-3}$.

\subsection{Stability \& Generalization}
\begin{figure}
\begin{minipage}{.5\linewidth}
\centering
\subfloat[]{\label{main:a}
    \begin{tikzpicture}
\begin{axis}[xlabel={Iteration},ylabel={Loss function},width=1\textwidth, height= 0.7\textwidth, ymode=log]
\addplot[color=black]
table[x index=0,y index=1,col sep=comma] {data/FC_loss.dat};
%\addlegendentry{Fully-connected}
\end{axis}
\end{tikzpicture}
}
\end{minipage}%
\begin{minipage}{.5\linewidth}
\centering
\subfloat[]{\label{main:b}
\begin{tikzpicture}
\begin{axis}[xlabel={Iteration},ylabel={Loss function},width=1\textwidth, height= 0.7\textwidth, ymode=log]
\addplot[color=black]
table[x index=0,y index=1,col sep=comma] {data/Resnet_loss.dat};
%\addlegendentry{Resnet}
\end{axis}
\end{tikzpicture}
}
\end{minipage}\par\medskip
\centering
\subfloat[]{\label{main:c}
    \begin{tikzpicture}
\begin{axis}[xlabel={Iteration},ylabel={Loss function},width=0.5\textwidth, height= 0.35\textwidth, ymode=log]
\addplot[color=black]
table[x index=0,y index=1,col sep=comma] {data/NAIS_loss.dat};
%\addlegendentry{NAIS-Net}
\end{axis}
\end{tikzpicture}
}

\caption{Evolution of the loss function for different architecture: (a) Fully-Connected , (b) Resnet  and (c) NAIS-Net  over 20000 iterations, using the Adam optimizer with learning rate $10^{-3}$.}
\label{fig:main}
\end{figure}
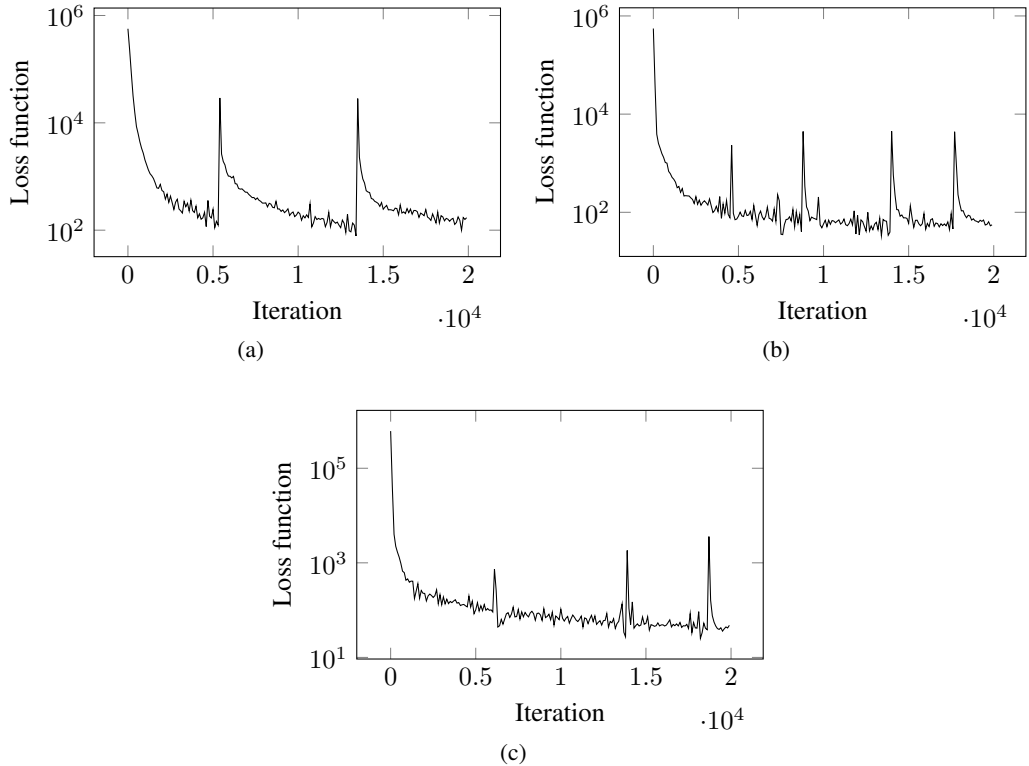

In Figure \ref{fig:main} we plot the evolution of the loss function for the three different cases: Figure \ref{main:a} Fully Connected, Figure \ref{main:b} Resnet and Figure \ref{main:c} NAIS-Net. We only plot the results after 20000 iterations. All three architectures eventually converge to a loss function of $10^{-4}$ and therefore very close to the global minimum.
It is clear from Figure \ref{fig:main} that NAIS-Net achieves a smoother reduction of the loss function. Note that for the fully connected case (Figure \ref{main:a}) it seems that the algorithm gets trapped in local minima and that it eventually manages to escape to a different and slightly better local minimum.
Resnet (Figure \ref{main:b}) appears to partly address this issue and NAIS-Net (Figure \ref{main:c}) has a much smoother profile.

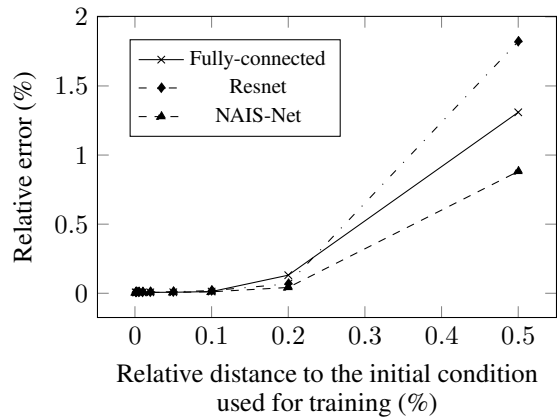
\begin{figure}[ht] 
\centering
\begin{tikzpicture}
\begin{axis}[xlabel={\makecell{Relative distance to the initial condition \\ used for training (\%)}},ylabel={Relative error (\%)},width=0.55\textwidth, height= 0.4\textwidth, legend style={at={(0.3,0.6)},anchor=south},   legend style={font=\fontsize{8}{5}\selectfont}]

\addplot[color=black,dash pattern=, mark=x]
table[x index=2,y index=3,col sep=comma] {data/archi.dat};

\addlegendentry{Fully-connected}

\addplot[color=black,dash pattern=on 3pt off 4pt on \the\pgflinewidth off 4pt, mark=diamond*]
table[x index=2,y index=4,col sep=comma] {data/archi.dat};

\addlegendentry{Resnet}

\addplot[color=black, dash pattern=on 3pt off 3pt, mark=triangle*]
table[x index=2,y index=5,col sep=comma] {data/archi.dat};

\addlegendentry{NAIS-Net}

\end{axis}
\end{tikzpicture}
  \caption{Average relative error of different architectures for different perturbations}
  \label{fig:generalisation}  
\end{figure}

To test the generalization properties of the different algorithms we perturbed the initial condition of the forward SDE. We then compared the relative error of the different architectures with respect to the closed form solution.
For this experiment we do not re-optimize the network parameters. Instead we use the optimal parameters of the unperturbed model. 
These experiments test how well the solution generalizes to different initial conditions.
We note that stability and generalization are linked (see \cite{bousquet2002stability}).
The results in Figure \ref{fig:generalisation} suggest that NAIS-Net computes a solution that generalizes better than the other two methods.
It is interesting to see that Resnet performs worse than the fully connected case.
A possible explanation for this is due to the forward stability problems of Resnet (see \cite{haber2019imexnet}).

\begin{figure}[t] 
  \begin{minipage}[b]{0.5\linewidth}
    \centering

\begin{tikzpicture}
\begin{axis}[xlabel={\makecell{t}},ylabel={$Y_{t}$},width=1\textwidth, height= 0.7\textwidth, legend style={at={(0.2,0.05)},anchor=south},   legend style={font=\fontsize{8}{5}\selectfont}]

%predict 1
\addplot[color=black,dash pattern=]
table[x index=0,y index=1,col sep=comma] {data/BS_results.dat};

\addlegendentry{Learned}

%predict 2
\addplot[color=black,dash pattern=, forget plot]
table[x index=0,y index=2,col sep=comma] {data/BS_results.dat};

%true 1
\addplot[color=black,dash pattern=on 3pt off 3pt]
table[x index=0,y index=3,col sep=comma] {data/BS_results.dat};

\addlegendentry{Exact}

%true 2
\addplot[color=black, dash pattern=on 3pt off 3pt]
table[x index=0,y index=4,col sep=comma] {data/BS_results.dat};

%scatter points
%\addplot [only marks,mark=square*] coordinates { (0,77.1) };
%\addplot [only marks,mark=square*] coordinates { (1,76.9) };
%\addplot [only marks,mark=square*] coordinates { (1,72.7) };

\end{axis}
\end{tikzpicture}

  \end{minipage}%%
  \begin{minipage}[b]{0.5\linewidth}
    \centering

\begin{tikzpicture}
\begin{axis}[xlabel={\makecell{t}},ylabel={Relative error},width=1\textwidth, height= 0.7\textwidth, legend style={at={(0.79,0.69)},anchor=south},   legend style={font=\fontsize{8}{5}\selectfont}]

\addplot[color=black,dash pattern=]
table[x index=0,y index=1,col sep=comma] {data/errors.dat};
\addlegendentry{Mean}

\addplot[color=black,dash pattern=on 3pt off 3pt]
table[x index=0,y index=2,col sep=comma] {data/errors.dat};
\addlegendentry{m + 2 std}

\end{axis}
\end{tikzpicture}

  \end{minipage}
  \caption{Results for Black-Scholes equation in 100 dimensions. Left panel shows the error in $Y_t$ for two paths, and the right panel shows the average error for all the paths.}
  \label{fig:BS_results}  
\end{figure}
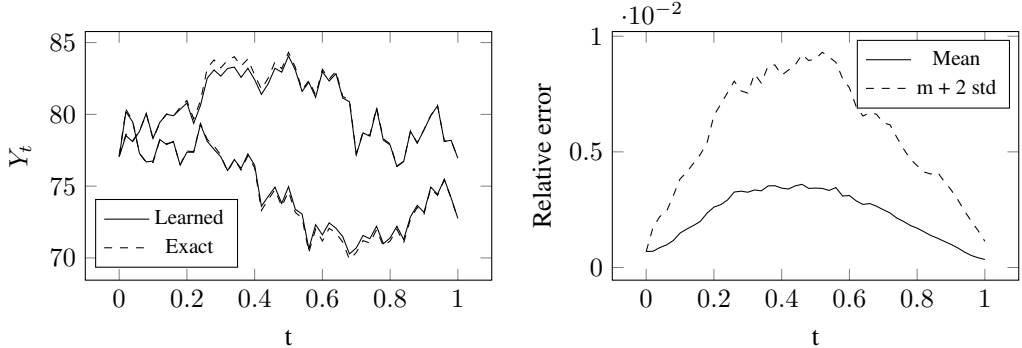

%Figure \ref{fig:main} shows the results obtained for different architecture, after 20000 iterations. The fully-connected network manages to solve the problem but suffers from stability. We notice that the  training  loss  is  very  unstable  with  multiple  peaks. Using Resnet helps to reduce the magnitude of peaks and using NAIS-Net helps to reduce both their magnitude and their number. This make the the network more stable during the training time but increases the computational time. Indeed, the fully-connected network requires 59 minutes to be trained, the Resnet needs 73 minutes and 123 minutes are required for the NAIS-Net.

%\subsection{Generalisation to different initial conditions}
%In the previous section, we suggest the use of some algorithms to increase the stability of the neural network. Some researches have linked stability and generalisation, using stability in machine learning algorithms to derive error bounds for generalisation (in particular \cite{bousquet2002stability}).
%We study the ability of networks based on different architecture to solve partial differential equations with a different initial condition than the one used for the training. To do so, we add increasing perturbations to the training initial condition and we study the average relative error between the approximation and the true solution.

%%% Generalisation figure
%%% End generalisation figure

\subsection{Solution times and speed-up}
\begin{figure}[ht] 
\centering
\begin{tikzpicture}
\begin{axis}[xlabel={Iteration},ylabel={Loss function},width=0.55\textwidth, height= 0.4\textwidth, ymode=log]
\addplot[color=black]
table[x index=0,y index=1,col sep=comma] {data/multilevel_loss.dat};
\end{axis}
\end{tikzpicture}
  \caption{Multilevel Monte Carlo }
  \label{fig:MLMC}  
\end{figure}
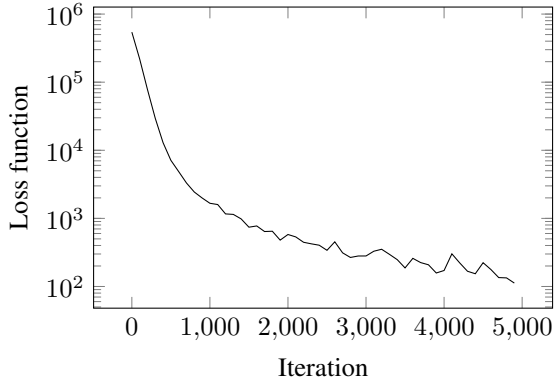

It is clear from the experiments of the previous section that NAIS-Net outperforms the other methods in terms of stability and generalization.  The solution times for the three different architectures are shown in Table \ref{table:solution times}.
NAIS-Net is the slowest (due to the extra projection step), but none of the methods are particularly fast.
In order to improve the solution times we implemented the multilevel technique described in Section \ref{sec:mlmc}. 
We generated samples using five different levels (2, 4, 8, 16 and 32 time steps). 
The convergence of the loss function shows a much more stable profile than in the single level case (see Figure \ref{fig:MLMC}).
The reduction in time and (at least for NAIS-Net) is an order of magnitude improvement over the single level method.
These initial encouraging results suggest that there is scope to investigate more advanced numerical discretization techniques for the system of FBSDEs.

\begin{table}[ht]
    \centering
  \begin{tabular}{|l||*{4}{c|}}
\hline
\backslashbox{Time (mins)}{Network}
&{Fully Connected}&{Resnet}&{NAIS-Net}
\\ \hline
Single-Level & 59& 73 & 123\\\hline
Multi-Level  & 6& 7 & 11\\\hline   
\end{tabular}
\caption{Solution times in minutes.}
\label{table:solution times}
\end{table}

%This reduces drastically the training time as shown in Figure \ref{fig:MLMC}. For the fully-connected network, we need 59 minutes to train the network. With the multi-level technique on the 5 defined levels, the training phase takes only 6 minutes, dividing the training time by almost 10.

\section{Conclusions}
In this paper, we reviewed the recent proposals for the solution of high dimensional PDEs using deep learning. The focus of the paper was on understanding the stability, robustness, and generalization of deep learning algorithms for this class of problems (as opposed to typical computer vision applications). We adopted the dynamical systems view and found that NAIS-Net, the nonautonomous architecture proposed in \cite{ciccone2018nais} is particularly well suited for this task. We also reduced the computational time associated with the optimization of neural networks by an order of magnitude. Our results suggest that deep learning can be applied to solving high dimensional PDEs (in the hundreds) and approximate solutions can be obtained within a few minutes using a standard computer with a GPU. The preliminary results suggest that more advanced discretization and sampling techniques are likely to yield even more improvements.

\section*{Acknowledgments}
This research was funded in part by JPMorgan Chase \& Co. Any views or opinions expressed herein are solely those of the authors listed,
and may differ from the views and opinions expressed by JPMorgan Chase \& Co. or its affiliates. 
This material is not a product of the Research Department of J.P. Morgan Securities LLC. 
This material does not constitute a solicitation or offer in any jurisdiction.
The work of the authors was partly  funded by Engineering \& Physical Sciences Research Council grants EP/M028240/1.

\clearpage

\small
\bibliographystyle{plainnat}
\bibliography{ref}
\end{document}